\documentclass[letterpaper, 10 pt, conference]{ieeeconf}

\IEEEoverridecommandlockouts

\usepackage{cite}
\usepackage{ulem}
\usepackage{acronym}
\usepackage{caption}
\usepackage{leftidx}
\usepackage{listings}
\usepackage{graphicx}
\usepackage{textcomp}
\usepackage{multirow}
\usepackage{booktabs}
\usepackage{colortbl}
\usepackage{subcaption}
\usepackage{algorithmic}
\usepackage{amsmath,amssymb,amsfonts}

\usepackage{booktabs}
\usepackage{array}
\usepackage{savesym}
\savesymbol{checkmark}
\usepackage{dingbat}
\usepackage{dingbat}
\usepackage{bbding}
\usepackage{colortbl}
\usepackage[table]{xcolor}
\usepackage{wrapfig}

\usepackage[utf8]{inputenc}
\usepackage{pgfplots}
\DeclareUnicodeCharacter{2212}{−}
\usepgfplotslibrary{groupplots,dateplot}
\usetikzlibrary{patterns,shapes.arrows}
\pgfplotsset{compat=newest}

\usepackage{hyperref}
\usepackage{cleveref}

\acrodef{TP}{True Positive}
\acrodef{FP}{False Positive}
\acrodef{FN}{False Negative}
\acrodef{VSLAM}{Visual SLAM}
\acrodef{STD}{Standard Deviation}
\acrodef{ROS}{Robot Operating System}
\acrodef{RMSE}{Root Mean Square Error}
\acrodef{ATE}{Absolute Trajectory Error}
\acrodef{RANSAC}{RANdom SAmple Consensus}
\acrodef{CNN}{Convolutional Neural Network}
\acrodef{LiDAR}{Light Detection And Ranging}
\acrodef{SLAM}{Simultaneous Localization and Mapping}

\definecolor{red}{HTML}{fd8f8f}
\definecolor{greend}{HTML}{57e377}
\definecolor{greenl}{HTML}{b8fb8a}
\definecolor{yellow}{HTML}{fefdb4}
\definecolor{orange}{HTML}{ffd5ab}

\colorlet{red}{red!50}
\colorlet{yellow}{yellow!50}
\colorlet{greenl}{greenl!50}
\colorlet{greend}{greend!50}
\colorlet{orange}{orange!50}

\title{\LARGE \bf BIM-Constrained Optimization for Accurate Localization and Deviation Correction in Construction Monitoring}

\author{
Asier Bikandi-Noya$^{1}$, Muhammad Shaheer$^{1}$, Hriday Bavle$^{2}$, Jayan Jevanesan$^{2}$, \\ Holger Voos$^{1}$, and Jose Luis Sanchez-Lopez$^{1}$ 
\thanks{$^{1}$Authors are with the Automation and Robotics Research Group, Interdisciplinary Centre for Security, Reliability, and Trust (SnT), University of Luxembourg, Luxembourg. Holger Voos is also associated with the Faculty of Science, Technology, and Medicine, University of Luxembourg, Luxembourg. \tt{\small{\{asier.bikandi, muhammad.shaheer, hermann.bavle, holger.voos, joseluis.sanchezlopez\}}@uni.lu}}
\thanks{$^{2}$ Authors are with GAMMA Tech.\tt{\small{\{hriday.bavle, 
jayan.jevanesan\}}@gamma-ar.com}}
\thanks{*
This work was partially funded by the Fonds National de la Recherche of Luxembourg (FNR) under the project C22/IS/17387634/DEUS, by a partnership between the SnT-UL and Gamma Tech.}%
\thanks{*
For the purpose of Open Access, and in fulfillment of the obligations arising from the grant agreement, the authors have applied a Creative Commons Attribution 4.0 International (CC BY 4.0) license to any Author Accepted Manuscript version arising from this submission.}
} 

\begin{document}

\maketitle
\thispagestyle{empty}
\pagestyle{empty}

\begin{abstract}
Augmented reality (AR) applications for construction monitoring rely on real-time environmental tracking to visualize architectural elements. However, construction sites present significant challenges for traditional tracking methods due to featureless surfaces, dynamic changes, and drift accumulation, leading to misalignment between digital models and the physical world. This paper proposes a BIM-aware drift correction method to address these challenges. Instead of relying solely on SLAM-based localization, we align ``as-built" detected planes from the real-world environment with ``as-planned" architectural planes in BIM. Our method performs robust plane matching and computes a transformation (TF) between SLAM (S) and BIM (B) origin frames using optimization techniques, minimizing drift over time. By incorporating BIM as prior structural knowledge, we can achieve improved long-term localization and enhanced AR visualization accuracy in noisy construction environments. The method is evaluated through real-world experiments, showing significant reductions in drift-induced errors and optimized alignment consistency. On average, our system achieves a reduction of $52.24\%$ in angular deviations and a reduction of $60.8\%$ in the distance error of the matched walls compared to the initial manual alignment by the user.

\end{abstract}
\section{Introduction}
\label{sec_intro}



The integration of Augmented Reality (AR) and Building Information Modeling (BIM) has improved construction monitoring by enabling real-time visualization of digital models on physical construction sites, allowing early detection of design issues and preventing costly errors \cite{yigitbas2023supporting}. Current AR applications typically rely on manual initial alignment, where users align a wall plane in the physical world with its corresponding BIM counterpart using a fixed geometric transformation. This process serves as a common baseline for AR-based construction applications, but it leads to drift accumulation over time and causes a progressive misalignment between the BIM model and the physical environment. 

\begin{figure}[t]
    \centering
    \includegraphics[width=1.0\columnwidth]{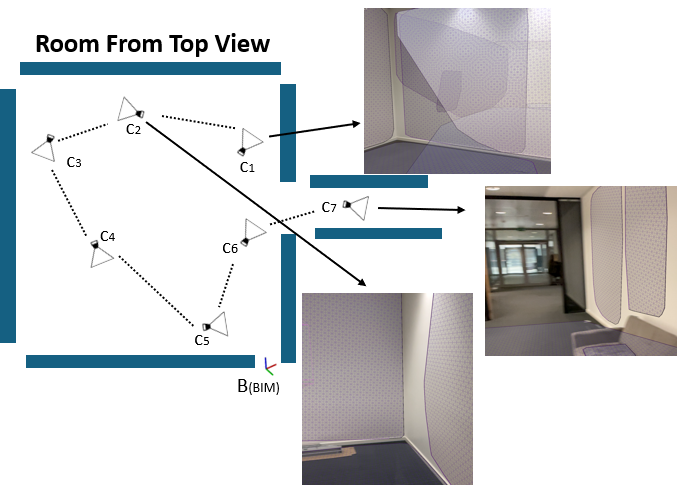}
    \caption{SLAM-Based Plane Detection and BIM Alignment for AR Visualization. From the top view, different camera poses are shown, and in the right side, real-world camera images are displayed with detected planes. 
    }
    \label{fig:AR_Global}
\end{figure}



Recent approaches have tried to optimize localization in construction environments. Blum et al. \cite{blum2022self}  developed a system that optimizes location for construction robots, nevertheless, it lacks real-time integration with structural models, crucial for accurate AR monitoring. Kuang et al. \cite{kuang2023ir} introduced IR-MCL for localization, but did not integrate BIM or any semantic context. In contrast, Real-Time Localization and Mapping with BIM \cite{shaheer2024real} effectively utilizes both BIM and semantic data for improved localization; however, it relies on LiDAR, which is computationally intensive and performs poorly on reflective surfaces, limiting its suitability for lightweight, real-time mobile AR applications.

In this context, RGB-D cameras, which combine color and depth information, have emerged as a popular alternative for AR-based applications due to their lightweight and affordability. While less precise than LiDAR, they enable real-time local depth-based alignment, making them ideal for portable devices like smartphones and tablets \cite{chidsin2021ar}.



\begin{figure*}[ht]
    \vspace{0.1pt}
    \centering
    \includegraphics[width=1.0\textwidth]{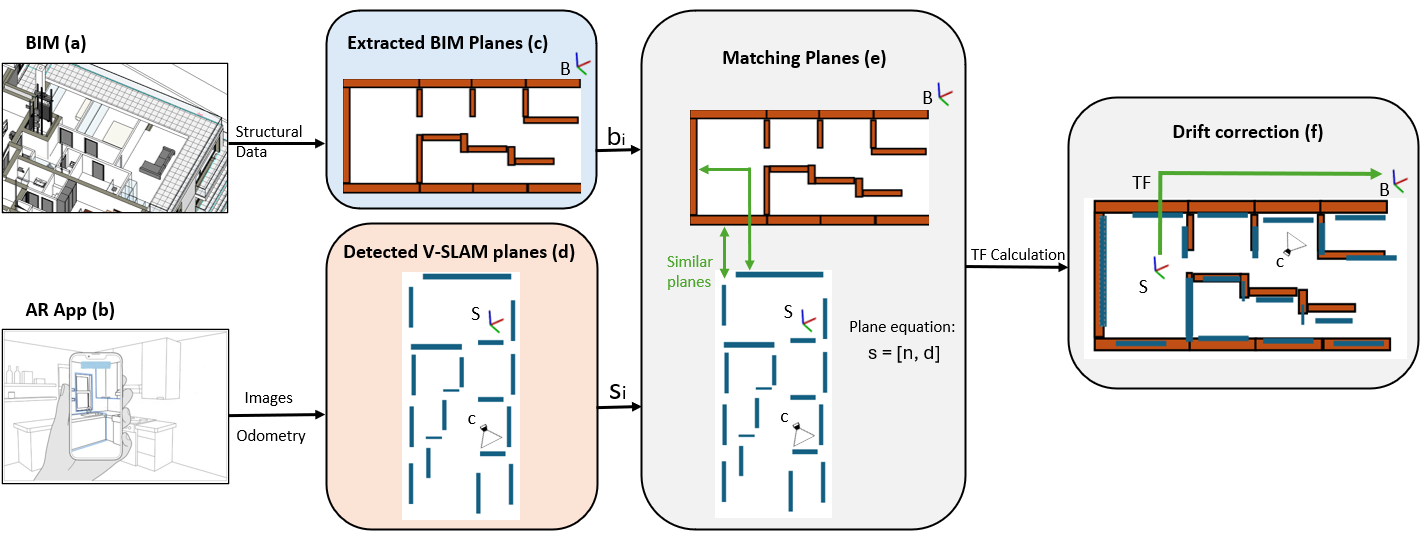}
    \caption{Overview of the system. Our approach extracts planar structures from SLAM and BIM models, applies robust plane matching, and computes a transformation (TF) using a Least Squares-based optimization method to minimize drift.}
    \label{fig_flowchart}
    \vspace{-2pt}
\end{figure*}


Although existing visual methods such as GS-SLAM \cite{yan2024gs} and RTG-SLAM \cite{peng2024rtg} offer valuable solutions to improve localization and map reconstruction, they do not use the semantic information retrieved throughout the mapping procedure. In contrast, vS-Graphs \cite{tourani2025vs} enhances scene comprehension by integrating semantic understanding, but, like the other methods, it does not incorporate BIM structural information for precise alignment and better contextualization of the environment. 


Construction sites present challenges for visual SLAM due to textureless surfaces and dynamic elements such as workers and vehicles, which disrupt feature tracking and cause drift accumulation \cite{sun2025nothing}, misaligning SLAM maps with BIM \cite{torres2023bim}. To address this, we propose a BIM-aware drift correction method for AR monitoring, using camera-based localization integrated with BIM constraints for better drift correction and long-term positioning. As shown in \Cref{fig:AR_Global}, our approach focuses on visualizing the elements of the construction site by aligning the extracted planes from different camera poses with BIM data, ensuring an accurate correspondence between physical and digital environments. Key contributions:

\begin{itemize}

\item BIM-aware localization for long-term positioning by integrating architectural data with real-time data obtained during application run.
\item A dynamic transformation estimation approach that continuously updates the alignment between BIM and SLAM reference frames, reducing drift over time.
\item Real-world evaluation of BIM-aware drift correction by detecting wall deviations in AR-based construction monitoring.

\end{itemize}

\section{SYSTEM OVERVIEW}
\label{sec_related}


Inspired by localization techniques such as \cite{shaheer2024real}, \cite{torres2023bim}, we propose a camera-based localization algorithm that incorporates BIM data to correct drift and optimize localization accuracy. The proposed system consists of three primary components:

\textbf{Environment Mapping.}  We map planar structures in real time using a camera (C), extracting relevant planes from the environment and relate them to S (see Figure \ref{fig_flowchart} (b), (d)).

\textbf{BIM Data Processing.} We extract and process structural data from BIM such as walls to match the detected planes from the real environment (see Figure \ref{fig_flowchart} (a), (c)).

\textbf{Plane Matching and Transformation Estimation.} We match the detected planes with the architectural ones and then compute the difference orientation between them. This allows us to calculate a transformation (TF) that aligns the reference frames, minimizing the rotation and translation errors of the detected walls (see Figure \ref{fig_flowchart} (e), (f)).

To ensure accurate alignment and minimize drift, our system uses a distance-based filtering technique to establish precise correspondences between the ``as-built''  and ``as-planned'' walls. An iterative optimization process then computes the optimal transformation to align the reference frames and dynamically correct the drift. By continuously integrating new detected planes, we refine the alignment over time, ensuring long-term stability and reducing misalignment.

\section{Methodology}
\label{sec_proposed}

\subsection{BIM-Based Plane Extraction and Matching}
\label{sec:semantic_matching} 
Plane detection is performed using a pre-trained model within a mobile application. Aligning the "as-built" and "as-planned" planes is crucial for reducing drift and enhancing localization accuracy. Our method addresses this by employing a BIM-constrained plane matching technique, ensuring geometric consistency between the two sets of planes. We apply a Mahalanobis distance-based filtering approach to robustly select matching candidates. This method effectively compares plane geometry, making it resilient to noisy or obstructed environments, and accounts for uncertainty in plane parameters to ensure the most probable matches, even under challenging conditions.


To improve efficiency, plane matching and TF estimation are performed only on keyframes with new or updated planes, avoiding unnecessary computations and maintaining accurate alignment. To determine if a BIM wall spans multiple rooms, we calculate the intersections between walls. If they overlap, we split the wall into separate sections, ensuring a one-to-one correspondence between the "as-built" and "as-planned" planes, thus improving matching accuracy and reducing alignment errors.


For a detected SLAM plane $s_i$ and a candidate BIM plane $b_i$, the Mahalanobis distance is computed as:



\begin{equation}
D_M(s_i, b_i) = (s_i - b_i)^T \Sigma^{-1} (s_i - b_i)
\end{equation}

where $\Sigma$ is the covariance matrix that captures the uncertainty in plane detection. A candidate plane $b_i^*$ is considered a match for $s_i$ if the distance is below a certain threshold $\tau$:

\begin{equation}
D_M(s_i, b_i^*) < \tau
\end{equation}

To reduce mismatches in feature-poor environments, we refine associations with geometric filters such as corner proximity, center alignment, and area to ensure structural consistency. 


\subsection{Optimization-Based Drift Correction}
\label{sec:drift_compensation}

\begin{figure}[h]
    \centering
    \includegraphics[width=0.8\columnwidth]{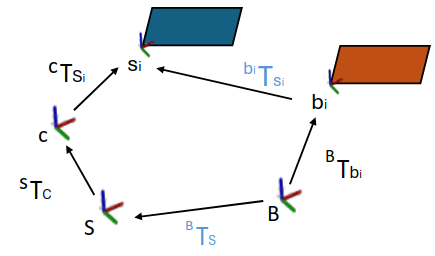}
    \caption{Defined reference frames for this work.}
    \label{fig:TF}
\end{figure}

As illustrated in \Cref{fig:TF}, our approach defines an error function that measures the transformation between matched planes to correct drift, as explained in \Cref{eq:transformation_error}.



\begin{equation}\label{eq:transformation_error}
e = {}^{b_i}T_{s_i} = \ominus {}^{B}T_{b_i} \oplus {}^{B}T_S \oplus {}^{S}T_C \oplus {}^{C}T_{s_i}
\end{equation}

The error, defined as ${}^{b_i}T_{s_i}$, represents the transformation from the "as-built" plane to the "as-planned" plane. The term ${}^{B}T_{b_i}$ denotes the BIM plane in B, while ${}^{B}T_S$ is the transformation between the origin frames. ${}^{S}T_C$ represents the pose of the camera relative to S, and ${}^{C}T_{s_i}$ is the plane detected by the camera.



With the objective of minimizing the error between the matched planes, the transformation ${}^{B}T_S$ is iteratively defined. This process is implemented using the Least Squares Method and a Gauss-Newton solver with a Dense Singular Value Decomposition (SVD) backend. This optimization is formulated as:

\begin{equation}
\arg\min_{R,t} \sum_{i=1}^{n} \| R s_i + t - b_i \|^2 + \| R n_s - n_b \|^2 
\end{equation}

We handle uncertainty in perpendicular translations by using at least three linearly independent planes, providing additional constraints to accurately estimate translation in all directions. 

In this work, the initial manual alignment serves as a baseline, simulating typical AR-based construction applications where users align the first BIM planes with physical walls. We match the first detected planes to their BIM counterparts using known wall IDs. This baseline enables comparison with dynamic transformation approaches, reflecting typical user interaction at the start of AR applications.


\section{Experimental Evaluation}
\label{sec_evaluation}

\begin{figure*}[t!]
\centering
\begin{subfigure}{0.48\textwidth}
 \centering
 \includegraphics[width=1\textwidth]{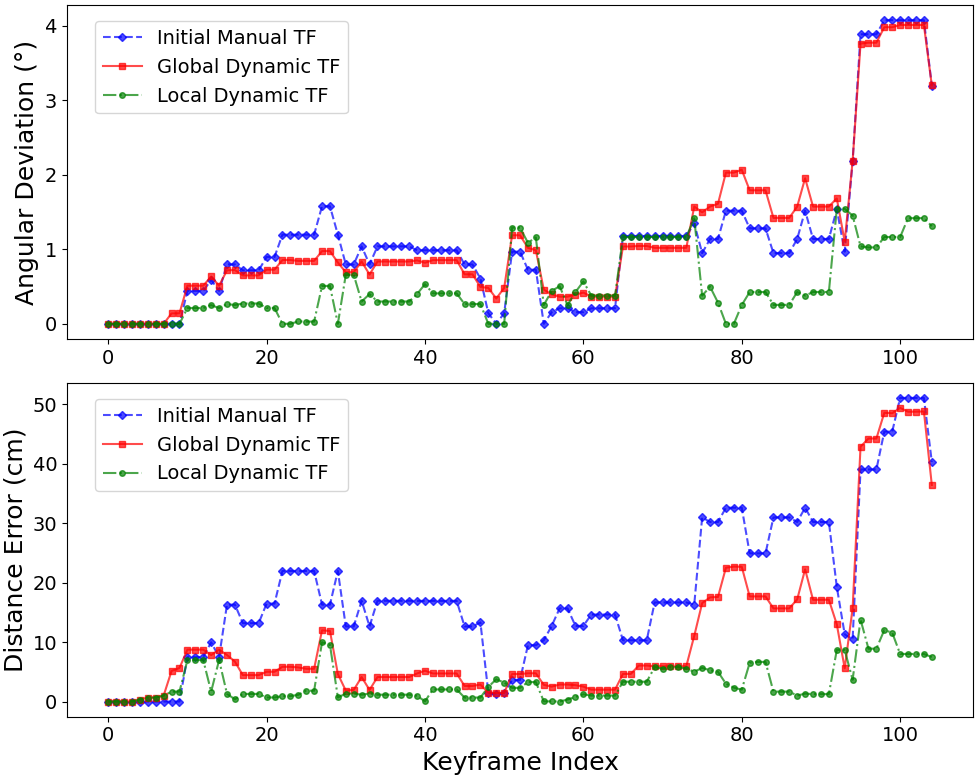}
 \caption{Office}
 \label{fig:Office}
\end{subfigure}
\begin{subfigure}{0.48\textwidth}
 \centering
 \includegraphics[width=1\textwidth]{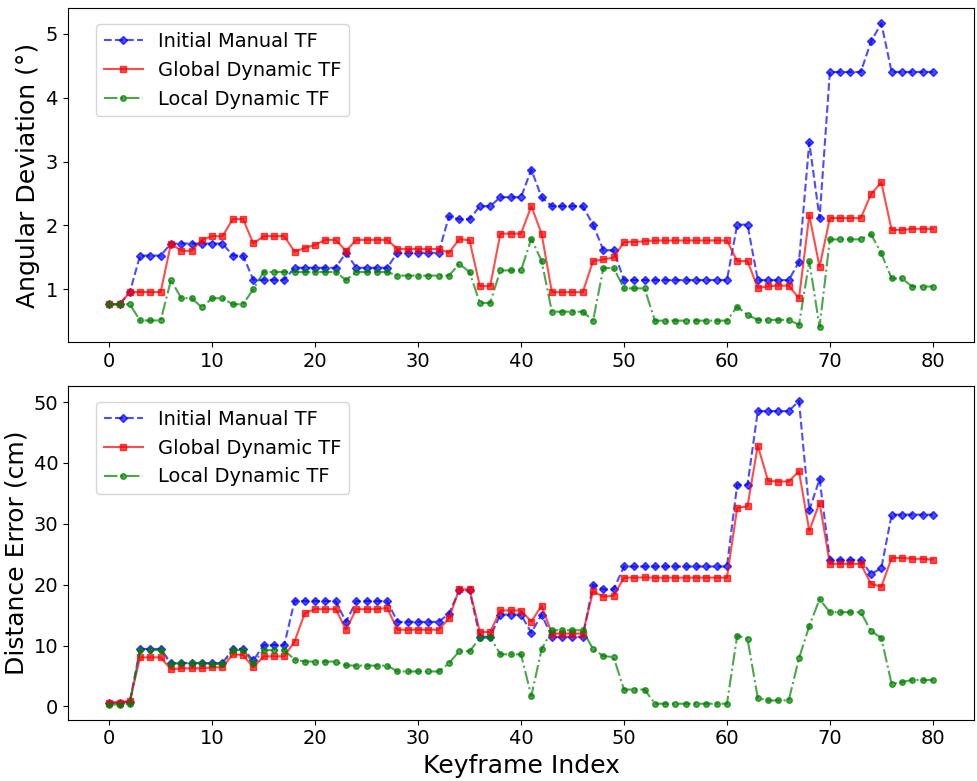}
 \caption{Construction Site}
 \label{fig:ConstructionSite}
\end{subfigure}

\caption{
Comparison of rotation and translation errors across two environments using three transformation estimation variants.
}
\label{fig:office_vs_construction}
\end{figure*}

\begin{figure*}[t!]
\centering
\begin{subfigure}{0.33\textwidth}
 \centering
 \includegraphics[width=1\textwidth]{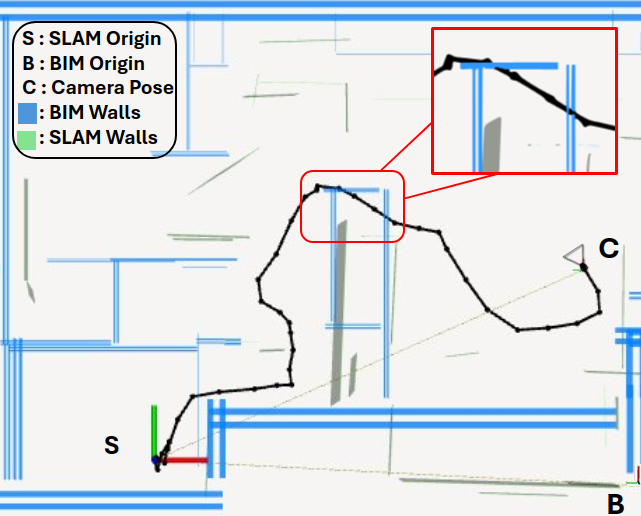}
 \caption{Initial Manual Transformation}
 \label{fig:}
\end{subfigure}
\hspace{2em}
\begin{subfigure}{0.33\textwidth}
 \centering
 \includegraphics[width=1\textwidth]{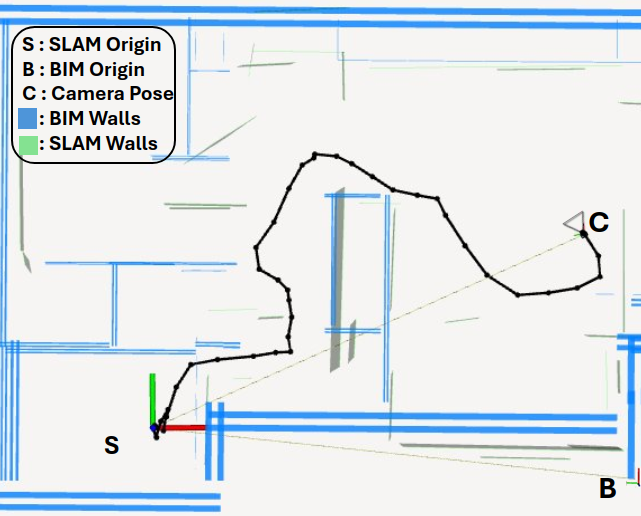}
 \caption{Local Transformation}
 \label{fig:ConstructionSite_2}
\end{subfigure}

\caption{
Comparison of camera pose trajectories. The left figure shows significant drift with the baseline method, colliding the walls in BIM, while the right one demonstrates the improved alignment with the Local Transformation approach. 
}
\label{fig:trajectories}
\end{figure*}

\subsection{Experimental Setup}

We evaluated our method in four real-world environments: two offices and two construction sites, using RGB-D camera data. The BIM data served as reference for the "as-built" planes. The evaluation metrics include the average distance error and the angular deviation of the matched planes, evaluating the alignment of SLAM-detected planes with BIM planes. We tested three variants based on the number of planes used for transformation estimation:

\textbf{Initial Manual Transformation.} Manual BIM alignment at the start with no further adjustment.

\textbf{Global Transformation.} Uses all matched planes for dynamic transformation.

\textbf{Local Transformation.} Uses nearby planes visible to the camera to calculate the dynamic transformation, gradually including more planes if necessary until a defined limit. 

\subsection{Results and Discussion}

Due to the strict page limit, we only present two of the four environments (1 office and 1 construction site) in \Cref{fig:office_vs_construction}. In both environments, the local transformation variant consistently outperforms the global and initial transformation in terms of distance error and angular deviation. This outcome was expected, as the graph plots errors of local matched planes, because the focus of this work is on visualizing the deviations of the planes observed around the camera in each keyframe.


The evaluation reveals an average reduction of $52.24\%$ in angular deviations and $60.8\%$ in distance error across the four environments when using the local transformation approach, compared to the initial manual one. Notably, as environmental noise increases or as the application runs for longer periods, the reduction in deviations becomes more significant, emphasizing the robustness of our approach in challenging conditions.

Although absolute trajectory errors cannot be computed due to the lack of a ground truth trajectory, \Cref{fig:trajectories} qualitatively illustrates the improvement of our method in a specific scenario of one of the construction sites in which the initial manual transformation method experiences significant drift, with the camera pose trajectory intersecting the BIM planes, an unrealistic result due to drift accumulation. In contrast, the Local Transformation configuration effectively mitigates this issue, ensuring a more accurate alignment throughout the trajectory.

\section{Conclusion}
\label{sec_conclusions}

This paper introduces a BIM-aware localization method for AR-based construction monitoring, minimizing drift and improving alignment between ``as-built'' and ``as-planned'' planes. Using a transformation to dynamically adjust the reference frame of BIM based on local matched planes, it outperforms traditional initial manual alignment by reducing orientation errors. The results demonstrate its effectiveness in real-time AR visualization, especially in noisy environments. A limitation of this work is the assumption that real-world planes have no construction deviations and align with BIM planes perfectly; future work will address this by incorporating uncertainties for such discrepancies and expanding to other construction elements.

\bibliographystyle{IEEEtran}
\bibliography{Bibliography}

\end{document}